%% file: main.tex
\documentclass[10pt,twocolumn,letterpaper]{article}

 \usepackage[pagenumbers]{iccv} 


\usepackage{url}
\usepackage{xcolor}

\usepackage{graphicx}
\usepackage{amsmath}
\usepackage{amssymb}
\usepackage{booktabs}
\usepackage{cuted}

\usepackage{multirow} 
\usepackage{array}

\newcommand {\otoprule}{\midrule [\heavyrulewidth]}
\newcolumntype {+}{ >{\global\let\currentrowstyle\relax}}
\newcolumntype {^}{ >{\currentrowstyle }}
 \newcommand {\rowstyle}[1]{\gdef\currentrowstyle{#1} %
 #1\ignorespaces
 }
\newcommand{\tabhead}{\rowstyle{\bfseries}}
\newcolumntype{K}[1]{>{\centering\arraybackslash}p{#1}}

\usepackage{xr}
\makeatletter

\newcommand*{\addFileDependency}[1]{
\typeout{(#1)}
\@addtofilelist{#1}
\IfFileExists{#1}{}{\typeout{No file #1.}}
}\makeatother

\newcommand*{\myexternaldocument}[1]{%
\externaldocument{#1}%
\addFileDependency{#1.tex}%
\addFileDependency{#1.aux}%
}
\myexternaldocument{supplementary}

\definecolor{iccvblue}{rgb}{0.21,0.49,0.74}
\usepackage[pagebackref,breaklinks,colorlinks,allcolors=iccvblue]{hyperref}

\def\paperID{11262} 
\def\confName{ICCV}
\def\confYear{2025}

\title{Efficient Unsupervised Shortcut Learning Detection and Mitigation in Transformers}

\author{Lukas Kuhn$^{*1,3,4}$, Sari Sadiya$^{*1}$, Jörg Schlötterer$^{2}$, Florian Buettner$^{1,3,4}$, Christin Seifert$^{2}$, Gemma Roig$^1$ 
\vspace{.3cm}\\
\parbox{\textwidth}{%
\centering
\normalsize
\begin{tabular}{@{}cc@{}}
$^{1}$Goethe Universität, Frankfurt &
$^{2}$Philipps-Universität, Marburg \\
$^{3}$German Cancer Research Center (DKFZ) &
$^{4}$German Cancer Consortium (DKTK)
\end{tabular}
}%
}

\begin{document}
\maketitle

\begin{abstract}
   Shortcut learning, i.e., a model's reliance on undesired features not directly relevant to the task, is a major challenge that severely limits the applications of machine learning algorithms, particularly when deploying them to assist in making sensitive decisions, such as in medical diagnostics.  In this work, we leverage recent advancements in machine learning to create an unsupervised framework that is capable of both detecting and mitigating shortcut learning in transformers. We validate our method on multiple datasets. Results demonstrate that our framework significantly improves both worst-group accuracy (samples misclassified due to shortcuts) and average accuracy, while minimizing human annotation effort. Moreover, we demonstrate that the detected shortcuts are meaningful and informative to human experts, and that our framework is computationally efficient, allowing it to be run on consumer hardware.
\end{abstract}

\section{Introduction}
\label{sec:intro}

\def\thefootnote{*}\footnotetext{These authors contributed equally to this work}\def\thefootnote{\arabic{footnote}}

Despite achieving performance comparable to that of human experts in many tasks, the deployment of deep neural networks still faces many challenges, especially in sensitive domains such as medical imaging~\cite{Rauker2023}. One major challenge is Shortcut Learning; a phenomenon where models exploit the presence of spurious features that coincidentally correlate with labels in the training data, despite not being relevant to the underlying relationship of interest. For instance, a review of hundreds of models for diagnosing COVID-19 from chest radiographs discovered that reported accuracies were often inflated by shortcut learning, with age being the spurious feature, and that none of the models can generalize to real world diagnostic settings~\cite{Roberts_2021}. Both the challenges of detecting when the model is learning shortcuts that rely on spuriously correlated (as opposed to core) features in the dataset, and of mitigating this shortcut learning, are active research topics that are also essential to the more general areas of fairness and bias mitigation in machine learning~\cite{Banerjee_2023}.
Despite recent progress, shortcut learning research still faces a number of challenges:
First, Shortcut mitigation approaches often require knowledge of which features are spurious. In addition, many methods require group annotations for a subset of the data, i.e., knowledge of both the class labels and the presence / absence of the spurious feature~\cite{liu21,Nauta_2022,kirichenko2023last}. However, this assumption rarely holds in real-world settings.   
    %
Secondly, in methods that do not require group annotations, the user often has no knowledge of the features were suppressed~\cite{zhang2021,Tiwari_2023}. However, use in sensitive domains requires the development of methods that allow users to retain control over the behavior of the model.  
Third, many current methods involve modifying the data and retraining the model~\cite{liu21,zhang2021,Nauta_2022,Asgari_2022}. These processes are highly computationally expensive. However, for a shortcut mitigation method to be widely adopted, it should support easy exploratory analysis of model behavior when suppressing different correlations.




In this paper, we introduce a framework for effective, efficient, and explainable shortcut detection and mitigation in transformers. Our approach leverages recent advancements in explainable AI, including prototype learning~\cite{Chen2019,Nauta_2021_CVPR} and Multi-Modal Large Language Models (MLLMs) concept identification~\cite{yang2023language}, to enable explainable unsupervised shortcut detection and mitigation.
The framework employs a multi-step process that analyzes model activations, detects prototypes in image patches, and uses MLLMs for prototype interpretation. The resulting interpretable shortcut detection component is then used to mitigate shortcuts during inference by selectively ablating image patches containing spurious features. Overall, our approach yields significant improvements in both worst group and average accuracy, offering a promising new direction in addressing shortcut learning. Finally, we conduct a user study to confirm that the prototype concepts discovered provide a meaningful insight into the spurious features present in the data. In summary, our contributions are the following:
\begin{itemize}

    \item We propose an end-to-end shortcut detection and mitigation framework to assist domain experts. Our framework is interpretable, interactive, and computationally efficient, requiring only accessible consumer-grade hardware.
    
    
    
    \item We show that the framework outperforms state-of-the-art techniques that require group annotation. Furthermore, the framework eliminates shortcuts from the stimuli directly, alleviating concerns regarding any remanence of shortcut learning \cite{Le_2023}.
    
    \item We conduct a user study to demonstrate that the shortcut detection component of the framework learns human-friendly prototype concepts that enable users to correctly identify dataset shortcuts, making our framework interpretable, in contrast to previous unsupervised methods.
    
\end{itemize}


\noindent To encourage further experimentation, all code and data necessary to replicate the experiments discussed in this paper are made publicly available\footnote{\href{https://github.com/Arsu-Lab/Shortcut-Detection-Mitigation-Transformers}{github.com/Arsu-Lab/Shortcut-Detection-Mitigation-Transformers}}.

\section{Related Work} \label{sec:relatedWork}

Despite recent advances in machine learning, including the rise of transformer-based architectures that surpass convolutional neural networks as the state-of-the-art, shortcut learning remains a persistent challenge~\cite{Ghosal_2022}. Many studies tackling shortcut learning assume an ideal scenario in which spurious features are known and group annotations are available~\cite{liu21, Nauta_2022, kirichenko2023last}. However, recent evidence suggests that even widely used datasets, such as ImageNet, contain previously unidentified shortcuts~\cite{Li_2023}. With this in mind, we explicitly tackled both shortcut detection and mitigation, without presuming any prior knowledge.


\subsection{Shortcut Detection}

Many approaches to shortcut detection focus on gradient-based methods, such as Grad-CAM \cite{Selvaraju_2019}, to identify spurious features. For example, DeGrave \etal. \cite{degrave_ai_2021} used gradients to create saliency maps, highlighting pixel importance across different COVID-19 X-ray images, and Zech et al. \cite{Zech_2018} applied activation maps to identify areas in chest radiographs that strongly influence model decisions. Although these methods have shown some success, they have been criticized for failing to provide the necessary insights to help users reliably identify spurious features~\cite{adebayo2022posthocexplanationsineffective}.

An alternative approach suggested by Mueller \etal.~\cite{müller2023shortcutdetectionvariationalautoencoders} uses Variational Autoencoders to detect latent space dimensions with high label predictiveness and generate a set of images that differ in a single image attribute. The user then manually inspects these image sets to detect the attribute being manipulated and makes a decision as to whether it is a core or spurious feature.


\subsection{Shortcut Mitigation}


Under the ideal scenario, in which the spurious features are known and group annotation (shortcut presence and label) is available for at least a subset of the data, it is possible to edit model representations to eliminate reliance on encoded spurious attributes. For instance, Kirichenko \etal \cite{kirichenko2023last} proposed Deep Feature Reweighting (DFR), which utilizes last layer retraining with a group-wise balanced dataset to effectively deactivate neurons that encode the spuriously correlated attribute. However, follow up work demonstrated that this may not be sufficient, as most neurons encode a combination of core and spurious attributes \cite{Le_2023}, suggesting that more exhaustive techniques are necessary to crowbar spurious and core attributes apart. For instance, Liu \etal \cite{liu21} uses extensive retraining while up-weighing samples from the worst performing group. Similarly, Idrissi et al. \cite{idrissi2022} proposed SUBG (subsampling groups), which reduces all groups to the size of the smallest group by randomly subsampling examples from larger groups, effectively removing group imbalance during training. While these methods achieve very high performance, their reliance on group annotations hinders their deployment in real-world settings.      

More recently, various approaches were developed that alleviate this dependency on group annotations \cite{zhang2021,Asgari_2022,Tiwari_2023}. Asgari \etal used saliency maps to identify which parts of the image the classifier depended on to reach its decisions and masked these parts in future training to encourage model exploration, thereby mitigating model reliance on any one specific attribute \cite{Asgari_2022}. Tiwari \etal \cite{Tiwari_2023} rely on a simple observation: spurious attributes are easier to learn relative to core attributes and are learned by early model layers \cite{Tiwari_2023}. Therefore, by applying a loss to early layers that penalizes high classification accuracy, the model is encouraged to forget the simple spurious attributes. Similarly, Zhang \etal \cite{zhang2021} used the image labels to eliminate representation-space variability between samples belonging to the same class that resulted in sample misclassification. While these methods improve performance, they do not provide insights towards understanding what the model relies on. Moreover, considering the task-dependent nature of the status of an attribute as spurious or core, blind feature suppression might be detrimental to overall task performance.  

\subsection{Interpretable AI}
Similarly to ~\cite{müller2023shortcutdetectionvariationalautoencoders}, our approach also utilizes clustering in the representation space to detect images containing shortcuts. However, instead of using generative vision models, our work builds on `Prototype learning` approaches such as~\cite{Chen2019,Nauta_2021_CVPR}. These methods focus on learning human interpretable concept representations that allow classifiers to make decisions that can be justified to the end user. In this vein, Yang et al. \cite{yang2023language} proposed an innovative method that leverages MLLMs to generate semantic prototype descriptions and detect `bottleneck concepts` that uniquely identify specific labels. 

\begin{figure*}[t]
  \centering
  \includegraphics[width=0.82\linewidth]{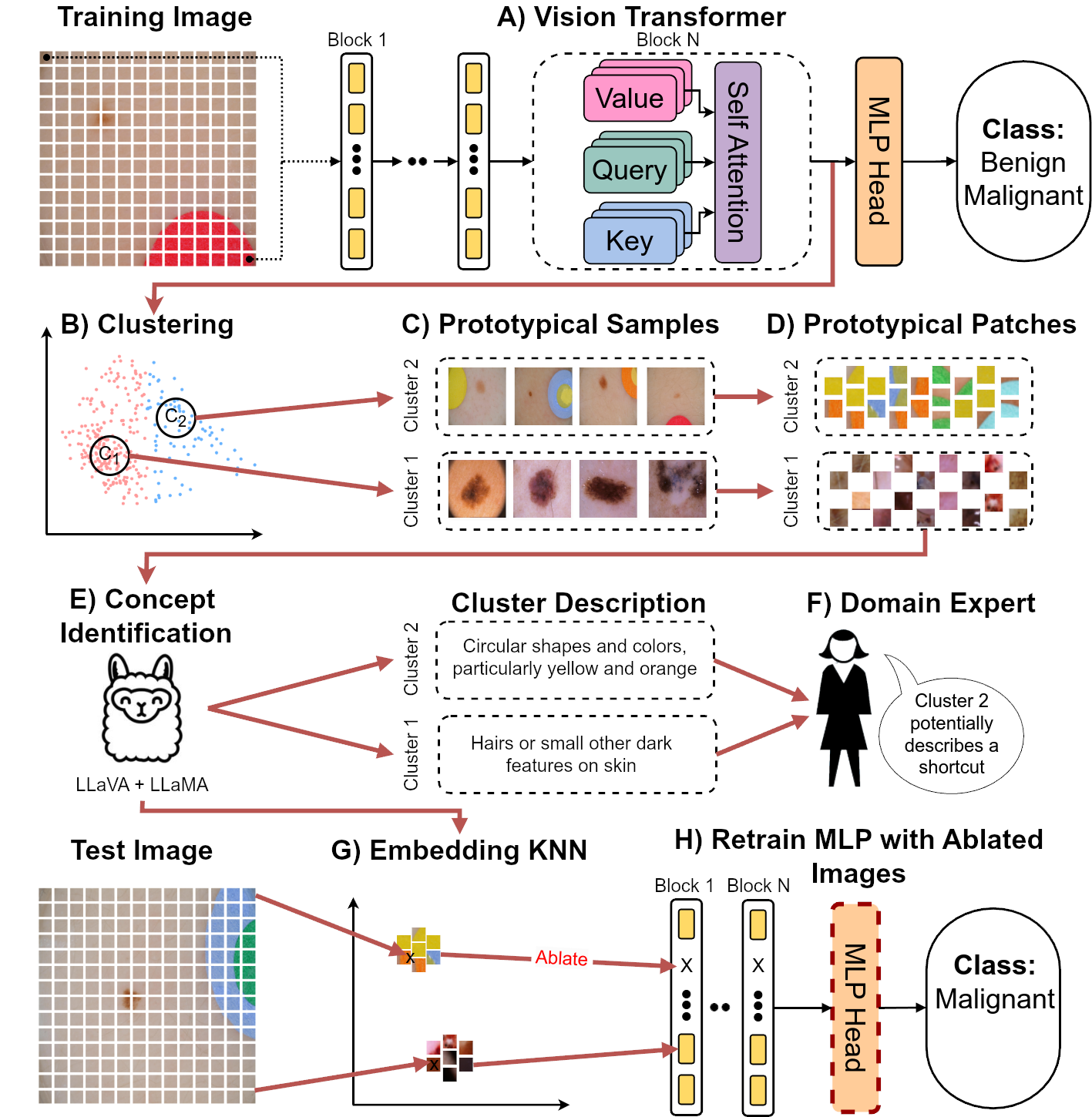}
   \caption{An overview of our approach. A) a vision transformer is fine-tuned to perform the classification task. B) the representations learned by the last block are used to cluster the images, key-space embeddings for each image patch (blue rectangles of the ViT) are saved to be used in step G. C) the most prototypical (closest to the cluster centroid) images are extracted. D) key-space distances to other patches in the same and other clusters are used to identify \textit{prototypical patches} for every cluster. E) multi-modal LLMs generate descriptions of prototypical patches in each cluster which are further distilled into unique \textit{cluster concepts}. F) domain expert in the loop uses these concepts to identify clusters containing spuriously correlated attributes. Alternatively, cluster label homogeneity, within/between cluster distance of the \textit{prototypical patches}, and Brier scores can be used for a completely unsupervised framework. G) Similarity to the key-space embeddings of the \textit{prototypical patches} are used to identify if spurious attributes are contained in validation image patches. H) Patches with high similarity to prototypical spurious patches are ablated, and the last linear layer is retrained to make decisions based on core features.}
   \vspace{-1em}
   \label{fig:wacv_meth}
\end{figure*}

\section{Approach} \label{sec:approach}

For our baseline, we used a ViT B-16 network pre-trained on ImageNet-1K that was fine-tuned on the training split of each of the datasets. We then execute our shortcut detection and mitigation pipeline using a holdout validation set.
The first step of our approach is to extract the patch activations from the final transformer block. These representations then undergo dimensionality reduction and clustering to identify salient patterns in the data. Within each cluster, we select representative samples based on their proximity to the cluster centroid, providing a representative subsample of the cluster's composition.

A key innovation in our approach is the identification of prototypical patches within these representative samples. We achieve this through a meticulous examination of token distances in the key space. This approach allows us to pinpoint the most influential patches within each image. We then employ a multimodal large language model (MLLM) to generate captions for these prototypical patches and comprehensive summaries of the patches within each cluster. These summaries allow a human expert to review - and if necessary intervene in - decisions made by the unsupervised pipeline, which employs simple yet effective heuristics to detect shortcut clusters. Our method then applies an ablation strategy that targets image tokens similar to those in the prototypical patch cluster identified as containing shortcuts. Finally, the last model layer is retrained to reinforce the reliance on core features during classification.

As demonstrated in Figure \ref{fig:wacv_meth}, the proposed approach is highly modular, consisting of a series of independent components. Although multiple alternatives for each component achieved comparable results, this set-up was ultimately chosen due to computational efficiency and simplicity.

\subsection{Clustering} \label{sec:app:clustering}
Neural networks embed images perceived to be similar (relatively to the task) into similar representations. We leverage this well-known observation for shortcut detection, as images containing spurious attributes often cluster in the embedding space \cite{Lapuschkin2019}. Therefore, the first step in our framework is clustering images in the representation space and identifying cluster prototypes (Figure \ref{fig:wacv_meth} A, B).
Let $X = {x_1, ..., x_n}$ be the set of images in our dataset. We define the embedding function $f: X \rightarrow \mathbb{R}^d$ that maps each image to a $d$-dimensional embedding space:
\begin{equation}
f(x_i) = \frac{1}{T} \sum_{t=1}^T h_L^t(x_i)
\end{equation}
where $h_L^t(x_i)$ is the embedding of the $t$-th token in the last layer $L$ for image $x_i$, and $T$ is the number of tokens (excluding the $<CLS>$ token).
We apply PCA to reduce the dimensionality from $d$ to $k$ (where $k=50$ in our experiments). We then apply unsupervised K-means clustering to these reduced embeddings. In contrast to Sohoni \etal \citep{Sohoni2020NoSL}, we do not apply overclustering, instead opting for the naive approach of clustering into two clusters. We verified the effectiveness of this approach against overclustering (with K selected by the silhouette method) and found no benefits, while also confirming clustering accuracy (see Appendix \ref{app:Clustering}).


\subsection{Prototypical Patch Identification} \label{sec:app:proto}
We focus on a representative subset of samples collected from each cluster by taking the $N$ nearest samples to the centroid (we report results for $N=20$). Given these samples, our next step is to identify \textit{prototypical patches}. These image patches should capture elements that are shared between images within each cluster, but are unique between different clusters (see Figure \ref{fig:wacv_meth} C, D, and E).

\subsubsection{Patch Similarity} \label{sec:app:proto:similarity}
Following Bolya \etal \cite{bolya2023tokenmergingvitfaster}, instead of using the intermediate and image model features, which can be overparameterized, we employ the keys learned in the last self-attention block of the ViT model for each patch. This is, we mean over the keys of the 12-heads to reduce dimensionality, obtaining a summary of the information in each image token. Let $K(p)$ be the mean of the attention heads' keys for patch $p$. To identify \textit{prototypical-patches} that are both unique for each cluster and important for the model's decision making, we compute a prototypicality score $P(p)$ for each patch, which is the Euclidean distance between each token and all tokens from other clusters:
\begin{equation}
P(p) = \frac{1}{|C_{other}|} \sum_{q \in C_{other}} ||K(p) - K(q)||_2
\end{equation}
where $C_{other}$ is the set of patches from all other clusters.
 Patches with high $P(p)$ scores are considered prototypical of their respective clusters (Figure \ref{fig:wacv_meth} E). We extract the top $M$ patches from the selected subset of each cluster for the following steps (we report results for $M=200$).

\subsection{Spurious Concept Identification} \label{sec:app:identification}

Given the patches we identified as prototypical in the previous step, we extract them from the pixel space and use them to generate human-understandable concepts. This is achieved using publicly available pre-trained MLLMs. Specifically, we first use LLaVA \cite{liu2023llava}, a model that bridges language and vision by learning a projection from the pretrained visual encoder of CLIP to Vicuna, an open source large language model \cite{vicuna2023}. We use LLaVA to extract text description from \textit{prototypical patches} for each cluster. The output is further refined to find the semantic contexts that best capture each cluster. This can be achieved efficiently using KL divergence to compute concept bottlenecks similar to previous work \cite{yang2023language}. However, we found that instructing a large language model to complete this refinement step generated superior performance. We tested three different LLMs and found excellent performance even when using the smallest LLaMA model \cite{touvron2023llama}. The text output allowed users to easily identify which cluster contains a shortcut (see the user study in Section \ref{sec:res}). While human input is not necessary, and this step can be skipped, the interpretability afforded by this step is crucial for real world applications.   

\subsubsection{Unsupervised Spurious Concept Identification}
Alternative to the prior step we developed a simple heuristic to identify a shortcut cluster, which we use in all our experiments. Although we believe human feedback is crucial for correct model behavior, this unsupervised approach could be useful when conducting cursory exploratory analysis. For unsupervised cluster selection, we used a combination of homogeneity and Brier score.
The cluster homogeneity score $h_c$ measures how much the cluster contains samples from a single class:
\begin{equation}
h_c = 1 - \frac{H(C|K)}{H(C)}
\end{equation}
where $H(C|K)$ is the conditional entropy of the class labels given the cluster assignments, and $H(C)$ is the entropy of the class labels.
The Brier score measures the accuracy of probabilistic predictions, defined as:
\begin{equation}
BS = \frac{1}{N} \sum_{i=1}^N (f_i - o_i)^2
\end{equation}
where $f_i$ is the predicted probability, $o_i$ is the actual outcome (0 or 1), and $N$ is the number of predictions.
The intuition behind this approach is the following: spuriously correlated attributes are, by definition, vastly overrepresented in a single class. Therefore, we expect that the cluster representing the spuriously correlated attributes will have high homogeneity. Similarly, the existence of the spurious feature is a strong signal of class membership. Therefore, as observed in previous research, the classifier will make confident predictions for images containing the spurious attribute \cite{Tiwari_2023}. Since these predictions are only correct for the dominant class, we calculate two Brier scores: one for the dominant ($bd_c$) and one for the nondominant ($bn_c$) class samples. We expect highly confident correct predictions for the dominant class samples leading to a low Brier score and highly confident incorrect predictions for the nondominant class samples (since it contains the spurious attribute but is not part of the spurious correlated class) leading to a high Brier score.
Hence, given the average homogeneity score $h_c$ and dominant Brier score $bd_c$ as well as non-dominant Brier score $bn_c$ per cluster, our unsupervised method picks the cluster that maximizes a combination of the three:
\begin{equation}
\underset{c \in C}{\mathrm{argmax}}, {\lambda_1 h_c+ \lambda_2 e^{-bd_c}+ \lambda_3 (1-e^{-bn_c})}
\end{equation}
Where the exponent of the Brier scores is used to map all elements to the same range. In our experiments, we weigh the three elements equally. As we discuss in Section \ref{sec:res:supatt}, a user study found that human judgments overwhelmingly aligned with the predictions of this unsupervised method.

\subsection{Spurious Feature Suppression} \label{sec:app:suppression}

Spurious feature suppression during inference is a two-step process. Given a test image, we first use the learned \textit{prototypical patches} to identify the parts of the image that contain spuriously correlated attributes. Then we use simple token ablation to eliminate these patches from the image, ensuring that the model does not rely on spurious attributes.    

\subsubsection{Spurious Feature Identification during Inference} \label{sec:app:identificationInference}
Detecting which patches, if any, in a test image contain spuriously correlated features is achieved using a supervised classification approach. Although simple approaches based on feature-space similarity to the previously computed \textit{prototypical patches} worked well, a K-Nearest Neighbors (KNN) classifier in the \textit{key-space} representation had superior performance and efficiency. Simply put, we construct a balanced training set using a bank of \textit{key vectors} from high $P(p)$ scoring patches as positive samples (class $1$) from the selected shortcut cluster, combined with an equal number of low $P(p)$ scoring key vectors from the non-shortcut cluster as negative samples (class $0$). Key embeddings are computed for all test image patches, and the trained KNN classifier determines which patches to ablate (Figure \ref{fig:wacv_meth} H).

Formally, let $K_s$ be the set of key vectors from high $P(p)$ scoring patches, $K_n$ be the set of key vectors from low $P(p)$ scoring patches, and $k_t$ be the key vector for a patch in the test image. We define the ablation criterion as:
\begin{equation}
\text{ablate}(p) = \begin{cases}
1, & \text{if } \text{KNN}_{\{K_s, K_n\}}(k_t) = 1 \\
0, & \text{otherwise}
\end{cases}
\end{equation}
where $\text{KNN}_{\{K_s, K_n\}}(k_t)$ represents the prediction of the KNN classifier trained on the balanced dataset of spurious ($K_s$) and non-spurious ($K_n$) key vectors. This classification-based approach provides a more robust framework compared to threshold-based methods, as it learns the decision boundary from both positive and negative examples. This approach generalizes well to all tested datasets.

\subsubsection{Spurious Feature Ablation} \label{sec:app:Ablation}
 Our framework takes advantage of the fact that the transformer architecture accepts an arbitrary number of patches by removing any patch that the previous step detected as containing spuriously correlated features from the tokenized representation. This ablation is done after the positional embedding; therefore, the embedding of all other image patches remains unaffected. Although more sophisticated approaches are viable, patch ablation is likely the simplest and most computationally efficient approach. As discussed in Section \ref{sec:res}, it also proved to be highly effective as images that do not contain spurious attributes are left largely unaffected (see Table \ref{tab:ablation}); therefore, unlike previous research, we avoid any adverse effects on accuracy of groups that do not contain the spurious attribute.  
 
\subsubsection{Last Layer Retraining} \label{sec:Retraining}



Having eliminated image patches that contain spurious features, the last step involves retraining to increase the models reliance on core features. We found that retraining the last dense transformer layer using the embedding of the images post-ablation achieved improved performance for little computational cost. Note that this last step is not analogous to the popular Deep Feature Reweighting approach for shortcut mitigation in convolutional neural networks, as we do not rely on group annotation \cite{kirichenko2023last}. Moreover, despite DFR being a powerful approach, researchers have found evidence that it is not sufficient to completely mitigate shortcut learning \cite{Le_2023}, instead arguing that only retraining with data from which spurious features were eliminated ensures that the model classifier head effectively forgets the spurious attributes completely \cite{Nauta_2022}. While editing training data is complex and computationally expensive, by utilizing the ablated images from the previous step of our approach (and leveraging the fact that transformers do not require a fixed number of input tokens), we can effectively ensure that the retrained model relies only on core features. 

\section{Experimental Setup} \label{sec:exp}

We benchmark our framework against popular shortcut mitigation methods that require group annotations: subsampling groups (SUBG) \cite{idrissi2022}, Deep Feature Reweighting \cite{kirichenko2023last}, and Just Train Twice \cite{liu21}. Following standard practice, we use worst and average group accuracy for evaluation (where groups are defined by label and shortcut presence). The \emph{Optuna} framework was used for hyperparameter optimization. 


\subsection{Models}
While the shortcut learning literature focuses mostly on popular convolutional neural networks, Transformer-based architectures have rapidly surpassed them as state-of-the-art in computer vision. Our work focuses on the ViT B-16 vision transformer architecture which remains ubiquitous and competitive in computer vision research \cite{dosovitskiy2020vit,Ghosal_2022}. The baseline model was pretrained using the common \textit{ILSVRC} subset of ImageNet\footnote{Pretrained weights were downloaded from the \textit{TorchVision} library}, and fine-tuning was done on each binary classification dataset individually. 

\subsection{Datasets}

We benchmark our approach on three different datasets. Similarly, to previous research \cite{Ghosal_2022} we found that vision transformers are not susceptible to many of the shortcuts found in popular datasets (for instance, worst group accuracy in the popular waterbird dataset was over $~90\%$, see Appendix \ref{app:celebA}). Therefore, we utilized newer, more challenging, datasets. Moreover, considering interpretability and computational accessibility are especially important in medical diagnostic tasks, we focus on medical datasets. Each experiment was repeated with three sequential seeds. 


\subsubsection{ISIC}
The International Skin Imaging Collaboration (ISIC) skin lesion dataset \cite{Codella2018ISIC} is a popular dataset in the shortcut learning literature \cite{Le_2023,Nauta_2022}. This dataset contains images of malignant and benign tumors (2000 training samples in each class). Of the benign samples, 1000 images contain colored bandages, whereas only 10 (artificially generated) are to be found in the malignant training samples, thus constituting a spuriously correlated attribute. The validation set is 10\% of the size of the train set with a similar group distribution. In addition, we generated 100 malignant tumor images in which we inserted bandages (see Appendix \ref{app:ISIC}). The worst group accuracy for this dataset refers to the ratio of correctly classified malignant samples with bandage.     

\subsubsection{Knee Radiographs}
The knee osteoarthritis radiograph dataset \cite{CHEN201984} contains knee joint X-ray images of 4796 participants suffering from various degrees of osteoarthritis. Due to ViT outperforming CNNs, unlike in previous work \cite{adebayo2022posthocexplanationsineffective} we used the more challenging task of classification of 'no' vs. 'moderate' osteoarthritis. Following previous work \cite{adebayo2022posthocexplanationsineffective,DeGrave2020.09.13.20193565} we added a hospital tag to 50\% and 2.5\% of the images from healthy and arthritic patients respectively. The training and validation sets contained 1000 and 200 samples per class, respectively (with the same group distribution). For testing, we generated 100 samples of each class with and without a spurious marker. The worst group in this dataset refers to moderate arthritis images with the added radiographic marker. 


\begin{figure}[t]
  \centering
  \includegraphics[width=0.95\linewidth]{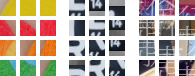}
   \caption{Prototypical patches from four samples selected by the framework. Left: The ISIC patches that were detected as prototypical of the shortcut class all contain the colored bandages. Middle: The knee radiographs prototypical shortcut patches mostly focused on the letters and numbers. Right: Imagenet-W prototypical shortcut patches mostly focus on the Chinese letters.}
   \label{fig:proto-patches}
   \vspace{-1em}
\end{figure}

\subsubsection{ImageNet-W}
Zhiheng \etal \cite{Li_2023} discovered that the \emph{carton} class in the popular ImageNet dataset \cite{ImageNet} is spuriously correlated with the presence of Chinese characters due to watermarks. We utilize this already existing shortcut in combination with the visually similar \emph{storage} image-net class, which does not contain any watermarks. We used 300 of the spuriously correlated carton class training images and a similar amount of the storage class images for training and 100 images from each for validation. For testing we used 100 images without watermarks of both the storage and carton class, and the same images with added watermarks (using code provided by Zhiheng \etal). The group with the worst accuracy was the artificially created watermarked storage samples.

\begin{table}[t]
\begin{center}
\begin{tabular}{+l^c^c^c}
\toprule \tabhead
                    & WGA (\%) $\uparrow$ & AGA (\%) $\uparrow$ & runtime$\downarrow$ \\\otoprule
                    & \multicolumn{3}{c}{\textsc{ISIC}}\\\cmidrule{2-4}    
 Baseline           & 51.7 $\pm$ 3.7  & 86.2$\pm$ 0.9   & n.a. \\
 SUBG                & 52.7 $\pm$ 3.7  & 74.3 $\pm$ 1.3  & 256.9 $\pm$ 16 \\
 DFR                & 59.3 $\pm$ 6.8  & 87.7 $\pm$ 1.4  & 27.3 $\pm$ 1 \\
 JTT                & 66.3 $\pm$ 4.1  & 82.7 $\pm$ 1.2  & 2886.3 $\pm$ 79 \\
 ASM (ours)      & \bf 74.7 $\pm$ 7.0  & \bf 88.7  $\pm$ 1.0  & 121.1 $\pm$ 9\\
  w/o retraining                 & 61.0 $\pm$ 2.4 & 87.3  $\pm$ 1.2   & 89.4 $\pm$ 2 \\\midrule 
                    & \multicolumn{3}{c}{\textsc{Knee Radiographs}}\\\cmidrule{2-4}        
 Baseline           & 37.3 $\pm$ 4.8  & 76.5  $\pm$ 1.3   & n.a. \\
 SUBG               & 47.7 $\pm$ 7.6  & 57.3 $\pm$ 3.7  & 248 $\pm$ 7 \\
 DFR                & 41.7 $\pm$ 7.8  & 77.8 $\pm$ 1.6   & 38.2 $\pm$ 1 \\
 JTT                & 40.0 $\pm$ 2.8  & 78.1 $\pm$ 0.2  & 1546.0 $\pm$ 46 \\
 ASM (ours)         & \bf 81.7 $\pm$ 6.3  & \bf 81.0  $\pm$ 0.9   & 99.8 $\pm$ 5\\
  w/o retraining                &  80.0 $\pm$ 8.6  &  80.7  $\pm$  1.0  & 82.9 $\pm$ 2 \\\midrule
                    & \multicolumn{3}{c}{\textsc{ImageNet-W}}\\\cmidrule{2-4}        
 Baseline           & 69.0 $\pm$ 4.2  & 91.2  $\pm$ 1.4 & n.a. \\ 
 SUBG               & 80.3 $\pm$ 3.3  & 90.8 $\pm$ 1.8  & 112.5 $\pm$ 4.2 \\
 DFR                & 80.3 $\pm$ 4.7  & 93.5 $\pm$ 1.4   & 48.1 $\pm$ 1\\
 JTT                & 72.0 $\pm$ 2.9  & 91.9 $\pm$ 0.6  & 488.5 $\pm$ 14 \\
 ASM (ours)       & \bf 87.0 $\pm$ 6.5  & \bf 95.3  $\pm$ 1.0   & 88.2 $\pm$ 3 \\
 w/o retraining                & 75.0 $\pm$ 2.4  & 93.1  $\pm$ 1.0   & 77.5 $\pm$ 2 
 \\\midrule 
 
\end{tabular}
\caption{The baseline is a pretrained \textit{Vit-B-16} fine-tuned on each dataset. Our ablation-based shortcut mitigation (ASM) approach (with and without last layer retraining) is benchmarked against subsampling groups (SUBG), Deep Feature Reweighting (DFR), Just Train Twice (JTT). We report Worst and average group accuracy (mean $\pm$ std) and total runtime seconds (including inference).}
\label{tab:res}
\vspace{-2em}
\end{center}
\end{table}

\section{Results} \label{sec:res}
The results of our experiments demonstrate that, compared to popular shortcut mitigation approaches, our framework significantly improves the worst group accuracy without sacrificing average classification performance (Table \ref{tab:res}).  

Our unsupervised ablation-based shortcut mitigation framework (ASM) outperforms popular methods that require group annotation both in terms of worst- and average-group accuracy. Moreover, an ablation experiment demonstrates that simply eliminating patches containing spurious features, without any retraining, already results in a significant performance increase.

\subsection{Computational Efficiency}
We designed our framework with computational efficiency in mind. We ran the unsupervised versions of all experiments on a standard 16GB RAM Apple MacBook Pro (M3). As can be observed in Table \ref{tab:res}, our approach provides significant boost to worst group and average accuracy in a matter of minutes. Hence, our approach is especially promising for end users looking to interactively explore their datasets and models to identify and eliminate shortcut learning. 

\subsection{Spurious Feature Ablation}

Examples of prototypical patches identified as encoding spurious attributes in each dataset can be found in Figure~\ref{fig:proto-patches}. Embedding similarity (in the key space) to these prototypical patches is used to detect and ablate patches containing spurious attributes. As can be observed in Table~\ref{tab:ablation}, images that do not contain spurious attributes were mostly unaffected by our method. These results, in conjunction with the ablation study results (where only patch ablation without retraining was used), indicate that our detection component correctly identifies spurious attributes in the stimuli.

\begin{table}[t]
\begin{center}
\begin{tabular}{+c^c^c^c}
\toprule \tabhead
                            & SP$ (\%) \uparrow$ & NS (\%) $\downarrow$ & \\\otoprule
\textsc{ISIC}               & 89.6 $\pm$ 6.2  & 5.0 $\pm$ 0.4   &  \\
\textsc{Knee Radiographs}   & 100.0 $\pm$ 0.0  & 0.0  $\pm$ 0.0   & \\
\textsc{ImageNet-W}         & 86.0 $\pm$ 2.7  & 8.5  $\pm$ 1.9   & \\\bottomrule
 
\end{tabular}
\caption{Percentage (mean $\pm$ std) of samples where we mask at least one token in the images in which the shortcut is present (SP) or no shortcut is present (NS).}
\label{tab:ablation}
\vspace{-2em}
\end{center}
\end{table}

\begin{table*}[ht]
\begin{tabular}{+c^c^p{6.8cm}^p{6.5cm}}
\toprule \tabhead
    & LLM           & Shortcuts  & No-Shortcuts \\\otoprule
\multirow{6}{*}{\rotatebox[origin=c]{90}{ISIC}}
    & LLaMA-70b       & Circular shapes and colors, particularly blue and yellow, rather than actual skin cancer features & The model may be relying on the presence of skin in the image \\
    & LLaMA-8b        & Blue and yellow circles on white backgrounds, potentially due to high contrast and simplicity & The model focuses on hairs or other small, dark features on the skin\\
    & Mixtral8x7b     & The presence of blue or yellow circles, which could be unrelated to skin cancer diagnosis & A potential shortcut of focusing on skin textures and features\\\midrule
\multirow{6}{*}{\rotatebox[origin=c]{90}{Knee Radiogr.}}
    & LLaMA-70b       & The presence of letters and numbers (e.g., "L", "R", "14") in the patches. & Prominent shadows or contrasts between light and dark areas\\
    & LLaMA-8b        &  Frequent occurrence of the letters "L" and "R" with the number "14"  & Bones or body parts, particularly those with medical significance or injuries\\
    & Mixtral8x7b     & Single letters with numbers (R14, L) on white or black backgrounds & A repeated pattern of bones or shadowy shapes may be a shortcut in this dataset\\\midrule
\multirow{6}{*}{\rotatebox[origin=c]{90}{ 
 $\quad$ImageNet-W}}
     & LLaMA-70b       & Chinese writing or characters on the boxes & The prevalence of wooden objects \\
    & LLaMA-8b        &  Asian-inspired characters, which may be due to the presence of Chinese writing on some objects & The model appears to focus on wooden boxes or chests with various objects or features\\
    & Mixtral8x7b     & The presence of Chinese writing as a shortcut to distinguish chest images from cardboard boxes & Wooden boxes and containers, with or without locks, often containing a white or gold object \\\midrule
\end{tabular}
\caption{Concept descriptions for shortcut and non-shortcut clusters generated by different MMLLMs. Human experts can easily identify shortcut features based on the generated text. For more information regarding the prompts and setup see appendix~\ref{app:llm}}
\label{tab:llms}
\vspace{-1em}
\end{table*}

\subsection{Spurious Attribute Detection} \label{sec:res:supatt}
For a shortcut detection framework to be effective, it is necessary to capture attribute concepts that are semantically meaningful. In fact, previous research found that most shortcut detection methods do not provide users with enough information to identify spurious attributes \cite{adebayo2022posthocexplanationsineffective}. Motivated by previous research, we conducted an analogous study. We used the patch descriptions generated by the multimodal LLaVA model \cite{liu2023llava}, and then further refined the results using a text-only LLM.


\noindent \textbf{User Study:}
For each dataset, we used LLaVA to generate descriptions given the prototypical patches of the cluster identified to most likely represent a shortcut. Then three LLMs were prompted to identify potential shortcuts based on the LLaVA outputs. We manually verified that the correct attribute is always present in the top three responses for each LLM. Participants were given a description of the task and asked to identify which response most likely describes a shortcut (for additional details see Appendix \ref{app:llm}).\\ 
\noindent \textbf{Results:}
50 participants took part in the study. Each was given the top three responses from LLaMA-70B, LLaMA-8B, and Mixtral8x7B and descriptions of the datasets (with no visual exemplars). On average participants identified the spurious attributes with accuracy significantly above chance ($\mu = 91.2\%$, $SD = 18.4\%$, chance level $=51.3\%$, one-tailed t-test statistics $t(49) = 15.33$, $p<0.0001$), demonstrating that the responses generated by the LLMs were indeed helpful for the end-user even without prior familiarity with the dataset, or even visual examples of the patch prototypes. Considering that domain experts using our framework will also have access to the visual prototypical samples, this accuracy can be perceived as a lower bound. Overall, these results indicate that explainable AI techniques that employ LLMs for concept guidance \cite{yang2023language} can be successfully extended for shortcut detection.

\section{Discussion}

As demonstrated in Section~\ref{sec:res}, our approach offers an efficient framework for detecting and mitigating shortcuts in transformers. Beyond performance, several design choices distinguish our method from recent research on shortcut learning: As detailed in Section \ref{sec:relatedWork},  previous unsupervised methods often focus on mitigating spurious shortcuts without explicitly identifying them \cite{Tiwari_2023,zhang2021,Asgari_2022}. In contrast, our design ensures that spurious attribute detection is explicit and can readily incorporate human input. This is achieved through the integration of prototype learning \cite{Chen2019,Nauta_2021_CVPR} and MLLM-guided concept identification \cite{byun2023}.

This is important for multiple reasons: First, surveys demonstrate that domain experts are reluctant to trust black-box models, regardless of model accuracy \cite{Chen_2022}. Second, the identification of shortcuts is task dependent, making human supervision crucial; for example, Chinese characters may be spurious attributes in object classification but core features in geographic location prediction. We hope that our work inspires further research that focuses on the role of human interpretability in the mitigation of shortcut learning. \\

Finally, as noted in previous research \cite{Ghosal_2022}, we found that transformers are resistant to many shortcuts that affect convolutional networks, such as in CelebA and Waterbirds (Appendix \ref{app:celebA}). Hence, our work is a needed step toward developing techniques and benchmarks that address the unique characteristics of shortcut learning in transformers.

\section{Conclusion and Future Work}

This paper presented a computationally efficient unsupervised shortcut detection and mitigation framework. Unlike previous work, our method has the advantage of generating explicit spuriously correlated attribute descriptions that can be easily understood and evaluated by the end user. Moreover, our method is computationally efficient and does not require editing the training data or retraining the model, and is therefore well suited to assist human experts looking to \textit{interactively} explore their data and model behavior.  

Despite encouraging results, our approach has several limitations which we look forward to addressing in future work. Specifically, recent research has demonstrated that eliminating one spurious attribute could enhance the classifier's reliance on other, previously hidden attributes~\cite{Li_2023}. We have encountered a similar situation when applying our approach to the CelebA dataset: Instead of the prototypical patches capturing gender, they focused on cloth color, jewelry, and the presence of neckties (see Appendix~\ref{app:celebA}). Although medical data are often acquired in controlled environments and are relatively clean, future research focusing on shortcut learning in complex datasets is necessary.

\section*{Acknowledgments}
This work was funded by the  Hessian.AI Connectom Networking and Innovation Fund 2024 and the European Union (ERC, TAIPO, 101088594 to F.B.) grant. Views and opinions expressed are those of the authors only and do not necessarily reflect those of the European Union or ERC. Neither the European Union nor the granting authority can be held responsible for them.

{
    \small
    \bibliographystyle{ieeenat_fullname}
    \bibliography{main}
}

\clearpage

\begin{strip}
  \centering
  {\Large \textbf{Supplementary Material: Efficient Unsupervised Shortcut Learning Detection and Mitigation in Transformers}}\\[0.5ex]
\end{strip}

\appendix

\section{ISIC Dataset} \label{app:ISIC}

To assess the classifier's performance on ISIC images of malignant tumors with colored bandages (representing the worst group performance), we manually added colored bandages to malignant tumor images from the validation set. This was done by cutting patches from unused training images using the background removal model \emph{tracer\_b7}, available as an API on Replicate. We obtained cutouts of colored patches, which were then layered onto malignant tumor samples using GIMP, varying the size, color, and location of the patches based on the training distribution.

\section{Knee Radiographs} \label{app:Knee}
Radiographic markers are frequently used to indicate the orientation and body part of the image. We obtained a cutout of an R (right body part) and L (left body part) marker from a hand x-ray image which we cutout with GIMP. We then automatically inserted the marker based on which knee (left or right) is visible in the image and varied in which corner (upper left and right as well as lower left and right) the marker is being added. We also added some slight rotation (between -5 and 5 degrees) to the added marker to introduce a more natural shortcut. 

This follows the methodology introduced by Adebayo \etal \cite{adebayo2022posthocexplanationsineffective} where they added a text ("MGH") as an artifical hospital tag on the image. Our approach occurs frequently in a variety of datasets which makes it even more natural.

\section{Commonly used datasets} \label{app:celebA}

Recent research suggests that Vision Transformers are quite robust against spurious correlations in commonly used datasets such as Waterbirds and CelebA. Ghosal et al. \cite{Ghosal_2022} finetuned a ViT B-16 on Waterbirds resulting in a 96.75\% average group accuracy and a 89.3\% worst group accuracy. Similarly, they finetuned a ViT B-16 on CelebA resulting in an average group accuracy of 97.4\% and a worst group accuracy of 94.1\%. 

We replicated their CelebA results by fine-tuning a ViT B-16 with the same hyperparameters as for all our other datasets, achieving above 90\% AGA and WGA. Running our shortcut detection and mitigation framework on this dataset also replicates Li et al.'s findings \cite{Li_2023} that eliminating one shortcut in this dataset will result in another being chosen by the model.

\begin{figure}[t]
  \includegraphics[width=\linewidth]{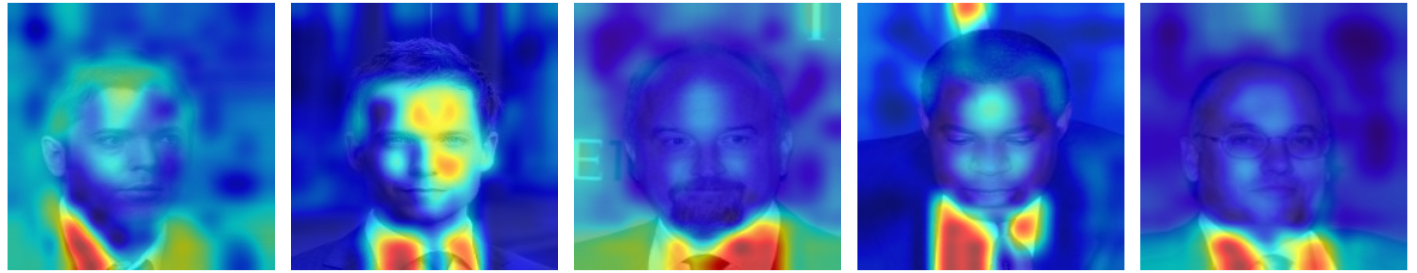}
   \caption{One of the many detected spurious correlations in the CelebA dataset are shirt collars for men.}
   \label{fig:celeba-collars}
\end{figure}

We detected a multitude of shortcuts which we were able to confirm as actual spurious correlations via the obtained labels (CelebA contains 40 labeled features including "necktie", "glasses" and "heavy-makeup"). A sample of a detected cluster can be seen in Figure~\ref{fig:celeba-collars}.

We again ran \emph{LLaMa-70b} to confirm the shortcuts in the prototypical patches and obtained the following summaries: 
\begin{itemize}
    \item "The model seems to focus on blonde hair, long hair,
ponytails, and beards/mustaches, which might be shortcuts for
identifying women or men."
    \item "The model appears to focus on facial
features like smiles, eyeshadow, lipstick, and moles, as well as
accessories like glasses, ties, and shirts"
\end{itemize}

\section{Clustering} \label{app:Clustering}
We calculated the average cluster accuracies for all datasets for all three sequential seeds (see Table \ref{tab:clusteraccuracies}). As described in Section 3.2.1, clustering results are refined using a patch similarity measure, ensuring robustness even if the initial clustering is imperfect. Although we lack ground-truth annotations for shortcut feature locations, manual inspection of 200 ISIC prototypical patches confirms that all patches in the shortcut cluster contain the expected spurious feature. 

\begin{table}[h!]
\begin{center}
\begin{tabular}{+c^c^c^c}
\toprule \tabhead
                      & Accuracy (\%) $\uparrow$ & \\\otoprule
\textsc{ISIC}   &  76.0 &  \\
\textsc{Knee Radiographs}   & 100.0 & \\
\textsc{ImageNet-W}   &  91.6 & \\
 
\end{tabular}
\caption{Average KNN clustering accuracy for all three sequential seeds, using two clusters.}
\label{tab:clusteraccuracies}
\vspace{-2em}
\end{center}
\end{table}

We also couldn't see any improvements in overclustering, as proposed by Sohoni et al. \citep{Sohoni2020NoSL} (see Table \ref{tab:overclustering}).

\begin{table}[h!]
\begin{center}
\begin{tabular}{+c^c^c^c}
\toprule \tabhead
                      & WGA$ (\%) \uparrow$ & AGA (\%) $\uparrow$ & \\\otoprule
\textsc{2 Clusters}   & 61.0 $\pm$ 2.4  & 87.3 $\pm$ 1.2   &  \\
\textsc{3 Clusters}   & 56.7 $\pm$ 3.4 &  86.2  $\pm$ 1.7   & \\
 
\end{tabular}
\caption{Worst and average group accuracy (mean and standard deviation) after shortcut mitigation with different clusters.}
\label{tab:overclustering}
\vspace{-2em}
\end{center}
\end{table}

\section{User Study}\label{app:llm}
We used the Replicate API to easily obtain results for multiple open-source LLMs. We decided on using three recently released open-source models with different parameter sizes:
\begin{itemize}
    \item LLaMa3-8b: The smallest open source LLaMa model with 8 billion parameters.
    \item Mixtral8x7b: Mixture of experts architecture with 13 billion parameters.
    \item LLaMa3-70b: The LLaMa model with 70 billion parameters.
\end{itemize}

We prompted all three models with the same prompt: 
"I extracted patches from images in my dataset where my model seems to focus on the most. I let an LLM caption these images for you. I am searching for potential shortcuts in the dataset. Can you identify one or more possible shortcuts in this dataset? Describe it in one sentence (only!) and pick the most significant. No other explanations are needed. Descriptions:" followed by the captions that we obtained via the \emph{LLaVa-13b} model. 
The \emph{LlaVa-13b} model was prompted with the prototypical patches and the text prompt "What is in this picture? Describe in a few words.".

The study was conducted using a google forms. The participants were prompted with the dataset description and asked to identify which of the three responses was likely describes a spuriously correlated attribute. Often there were multiple correct answers, hence chance performance was $51.3\%$. Note that we used responses only based on the cluster our unsupervised method identified as the one most likely to contain spurious correlations. Therefore, the results of the survey validate that LLMs are capable of generating concepts that distill the properties captured by the patch prototypes.

\end{document}